\documentclass[a4paper,fleqn]{cas-dc}

\usepackage[authoryear]{natbib}

\usepackage{color}
\usepackage{threeparttable}
\usepackage{amsmath}
\usepackage{fancyhdr}

\def\tsc#1{\csdef{#1}{\textsc{\lowercase{#1}}\xspace}}
\tsc{WGM}
\tsc{QE}
\tsc{EP}
\tsc{PMS}
\tsc{BEC}
\tsc{DE}

\begin{document}
	\let\printorcid\relax
	\let\WriteBookmarks\relax
	\def\floatpagepagefraction{1}
	\def\textpagefraction{.001}
	\shortauthors{Wenqi Ren et~al.}

	\title [mode = title]{
		\includegraphics[width=1.0\textwidth]{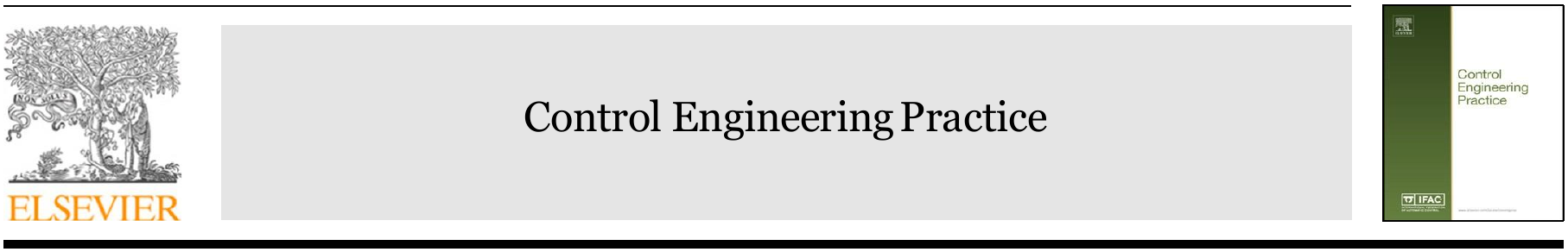}~\\
		\vspace{3pt}
		Towards Generalization on Real Domain for Single Image Dehazing via Meta-Learning}

	\newpage
	
	%
	%

	\author[1]{Wenqi Ren}
	\ead{wenqiren9801@163.com}
	
	
	\address[1]{Key Laboratory of Smart Manufacturing in Energy Chemical Process, Ministry of Education, East China University of Science and Technology, Shanghai 200237, China}

	\author[1,2]{Qiyu Sun}[%
	]
	\ead{qysun291@163.com}
	\address[2]{Shanghai Institute of Intelligent Science and Technology, Tongji University, Shanghai, China}
	
	%
	
	\author%
	[1]
	{Chaoqiang Zhao}
	\ead{zhaocqilc@gmail.com}
	
	\author[1]{Yang Tang}
	\cormark[1]
	\ead{yangtang@ecust.edu.cn}
	
	
	\cortext[cor1]{Corresponding author}
	

	\begin{abstract}
		Learning-based  image dehazing methods are essential to  assist  autonomous systems in enhancing reliability.  Due to the domain gap between synthetic and real domains, the internal information learned from synthesized images is usually sub-optimal in real domains, leading to severe performance drop of dehaizing models.  Driven by the  ability  on exploring internal information from a few  unseen-domain samples, meta-learning is commonly adopted to address this issue via test-time training, which is hyperparameter-sensitive and time-consuming. In contrast, we present a domain generalization framework based on meta-learning to dig out representative and discriminative internal properties of real hazy domains without test-time training. 
		To obtain representative domain-specific information, we attach two entities termed adaptation network and distance-aware aggregator to our dehazing network. The adaptation network assists in distilling domain-relevant information from a few hazy samples and caching it into a collection of features. The distance-aware aggregator strives to summarize the generated features and filter out misleading information for more representative internal properties. To enhance the discrimination of distilled internal information, we present a novel loss function called domain-relevant contrastive regularization, which  encourages  the internal features generated from the same domain more similar and that from diverse domains more distinct.  The  generated representative and discriminative features are regarded as some external variables of our dehazing network to regress a particular and powerful function for a given domain.  The extensive experiments on  real hazy datasets, such as RTTS and URHI, validate that our proposed method has superior generalization ability than the state-of-the-art competitors.

	\end{abstract}
	


	\maketitle

	\section{Introduction}
	The control and navigation of autonomous systems rely heavily on accurate visual perception \citep{ifqir2022fault,grof2022positioning,tang2021quadrotor},  in which clear images play an significant role \citep{zhao2022monovit,9803821,wu2022hybrid}. However, when affected by haze, images captured outdoors  usually  suffer from low contrast and poor visibility, resulting in severe performance degradations of perception models in autonomous systems.  Thus, it is significant to recover clear images from hazy counterparts, called image dehazing, for enhancing autonomous system reliability.

	Recently, benefit from the advances in deep learning, a great number of  learning-based image dehazing methods are proposed, which  provides  a solution to  promote the robustness of autonomous systems in foggy days \citep{zhang2020autonomous,tang2022}. 
	Nevertheless, the optimization of these dehazing models requires a large quantity of paired hazy/clear images, which  are laborious and expensive to be collected in practice. 
	To address this issue, numerous synthetic datasets \citep{resideli,li2020learning,sakaridis2018semantic,zheng2021ultra} are spwaned, where hazy images are synthesized from clear counterparts. However, these synthesized hazy images are biased to depict  real scenarios, leading to the domain gap between synthetic and real samples.   
	Thus,  the dehazing models learned from these synthetic samples  \citep{resideli,li2020learning,sakaridis2018semantic,zheng2021ultra}  frequently suffer from severe performance drop on  real images, as the internal information learned from synthetic domains is usually sub-optimal in real domains.

	Meta-learning  is capable of  digging out  internal information from a few samples of target domains,  so that the model only trained on source domains can quickly adapt to target domains \citep{zhang2020autonomous,sun2022learn}. Therefore, applying meta-learning to single image dehazing will be a promising subject to improve the generalization ability  on real domains.  Meta-learning can be categorized into metric-based, optimization-based and model-based approaches \citep{huisman2021survey,chen2022meta}, where optimization-based techniques \citep{finn2017model,liu2019self,sun2020test} are usually employed in image restoration \citep{soh2020meta,chi2021test} to explore the internal information of  new scenarios.  
	Nonetheless, these methods \citep{soh2020meta,chi2021test} depend heavily on test-time training, which   requires carefully designed hyperparameters (i.e., iteration steps)  for unseen test samples \citep{gao2022matters} and contributes to additional computational consumptions \citep{huisman2021survey}, which hinders the real-time application of autonomous systems \citep{kaleli2020development,zhang2022online,van2022optimal}. Different from these  researches \citep{soh2020meta,chi2021test}, we seek to deal with domain generalization by leveraging model-based meta-learning methods \citep{huisman2021survey,chen2022meta,garnelo2018conditional,zhang2021adaptive,ye2022contrastive}, epecifically adaptive risk minimization (ARM) \citep{zhang2021adaptive}, which  can effectively avoid test-time training and enable internal learning \citep{chi2021test,zhang2019hyperspectral} on real domains with a few real-world hazy samples. 
	
	To capture internal information of a given domain, \cite{zhang2021adaptive} add two entities termed adaptation network and average aggregator to the prediction network. The adaptation network assists in distilling internal information from a few samples and caching it into a collection of features. The average aggregator  strives to summarize the generated features to grasp domain-specific internal properties.  The summarized features are regarded as some external variables of the prediction network to regress a particular function for the given domain.  Intuitively, the more representative and discriminative the extracted domain-specific information is,  the more capable the regressed prediction function is to cope with samples in the given domain. However, directly employing ARM \citep{zhang2021adaptive} into image dehazing may cause some limitations.  Firstly, the adaptation network is made of vanilla convolution layers, which may lead to unrealiable internal information, as the inputs are generally covered by haze in image dehazing. Secondly, the average aggregator treats each sample equally and makes the external variables vulnerable to  outliers, since the features encoded from outliers usually fail to grasp representative domain-specific information.  Thirdly, the adaptation network and the aggregator only focus on intra-domain information  and leave inter-domain information unused,  leading to discriminative domain-specific information failing to be captured sufficiently.  
	
	Targeting at the first issue, we embed context-gate convolution (CG-Conv) layers \citep{lin2020context} into the adaptation network, which  enhances the reliability of features by fusing context information from entire images.  Aiming at the second challenge, we propose a non-parameter distance-aware aggregator to suppress the misleading information of outliers, according to the discovery that  features of outliers are usually located far away from that of normal samples. The proposed aggregator reweights  the internal features of diverse samples  according to their relative distances, so that the misleading information from outliers is weaken and the representative information of normal samples is enhanced.  To address  the third dilemma,  intra- and inter-domain information is incorporated and  a domain-relevant contrastive regularization is presented, which attempts to make the internal information distilled from the same domain more similar and that from diverse domains more distinct.  By embedding the domain-relevant contrastive regularization into our framework, the intra-domain homogeneity and inter-domain heterogeneity can be grasped, which further enhances the representativeness and discrimination  of external variables.  Comprehensive experiments are conducted to demonstrate the generalization ability of our dehazing model on real domains.

In summary, the main contributions are listed as follows: 
\begin{itemize}
	\item A domain generalization framework  is proposed for single image dehazing, which can generalize to real hazy images effectively without test-time training.
	\item A  distance-aware aggregator is presented to capture more representative information of normal samples and suppress the misleading one of outliers. 
	\item A domain-relevant contrastive regularization is presented, which facilitates the regressed dehazing function  to capture more discriminative internal information of domains.
	\item The experiments on real hazy images demonstrate that our proposed framework is superior to the state-of-the-art learning-based dehazing methods  \citep{yang2022self,guo2022image}.
\end{itemize}

The rest of this paper is organized as follows. Section \ref{section2} reviews some leading-edge studies related to both single image dehazing and meta-learning in image restoration. Section \ref{section3}  gives a detailed overview of our dehazing framework. Section \ref{section4} illustrates the implementation details and  experimental results. Section \ref{section5} summarizes this paper and discusses  the future work. 

\section{Related Work}
\label{section2}

\subsection{Learning-based Single Image Dehazing}
The past decade has witnessed the emergency of a large number of learning-based image dehazing algorithms. 
\cite{li2017aod}  predict  transmission maps  and atmospheric lights jointly in a unified CNN architecture and generates haze-free images through a variant of the atmospheric scattering model \citep{mccartney1976optics,narasimhan2000chromatic,narasimhan2002vision}.  More learning-based methods \citep{liu2019griddehazenet,dong2020multi,qin2020ffa,wu2021contrastive,guo2022image} tend to generate haze-free images directly, as estimating intermediate variables (i.e., transmission maps) may give rise to cumulative errors \citep{cai2016dehazenet,ren2016single,zhang2018densely,lee2020cnn}.  \cite{liu2019griddehazenet} abandon the estimation of transmission maps and design a end-to-end CNN network to conduct dehazing.  This method \citep{liu2019griddehazenet} fails to leverage features from different scales, which drives \cite{dong2020multi} to  dig out the correlations of multi-level features. Taking the importance of diverse features and pixels into account, \cite{qin2020ffa} employ the channel attention and the pixel attention to treat each feature  and each pixel unequally.  To balance the performance and computational costs, \cite{wu2021contrastive} devise a compact network architecture and introduces contrastive learning to suppress unexpected predictions. Driven by the ability of Transformer in modeling  long-range feature dependences,  \cite{guo2022image} integrate Transformer and CNN for single image dehazing.

Learning-based approaches  require a great deal of paired data  to optimize their models, which spawns various synthetic datasets \citep{resideli,li2020learning,sakaridis2018semantic,zheng2021ultra}. These hazy images with diverse densities  are synthesized from clear ones  by adjusting scattering coefficient and atmospheric light via the atmospheric scattering model \citep{mccartney1976optics,narasimhan2000chromatic,narasimhan2002vision}. Nevertheless, due to the inaccurate estimation of  depths and the complexity of real imaging mechanism, the synthetic hazy images fail to depict real-world hazy scenes reliably, contributing to the domain gap between synthetic and real-world hazy images. Although the model trained on synthetic datasets  \citep{resideli,li2020learning,sakaridis2018semantic,zheng2021ultra} can grasp the internal information  of synthetic domains, it frequently suffers from a performance drop on the real hazy images, as the features learned from synthetic domains are sub-optimal in real domains due to the domain gap.

\begin{figure*}[!t]
	\centering
	\includegraphics[width=\linewidth]{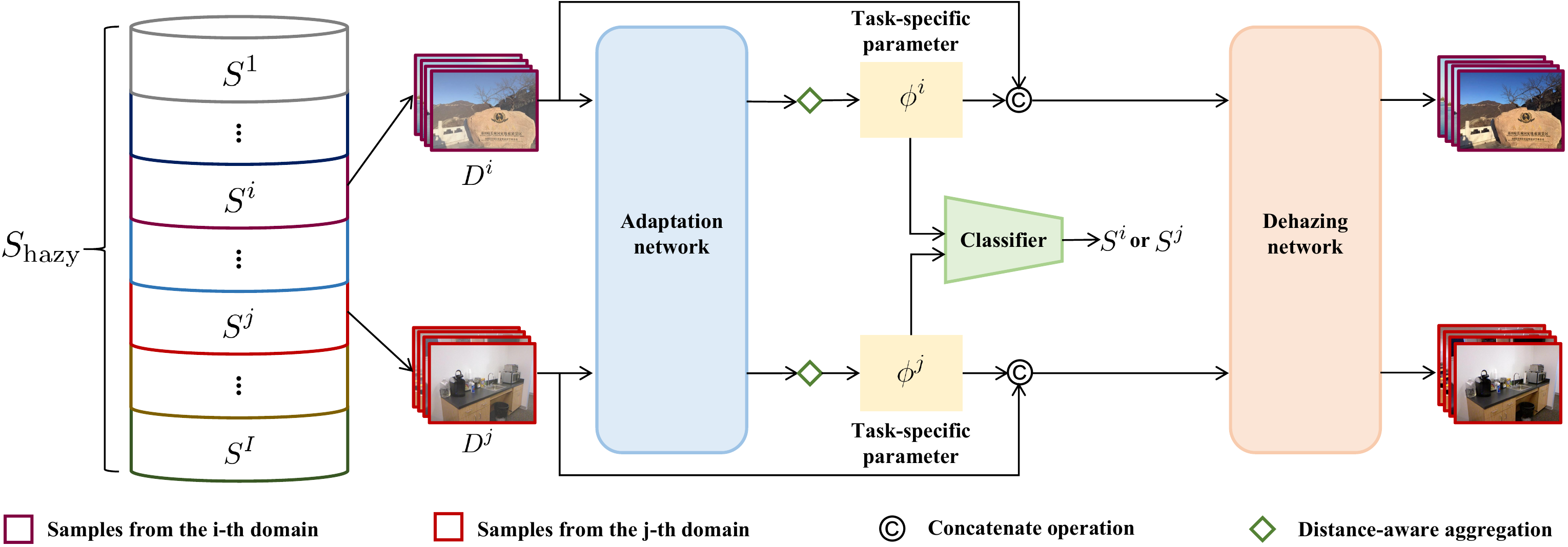}
	\caption{The overview of the proposed framework.}
	\label{fig:framework1}
\end{figure*}

\begin{table}
	\centering
	\caption{The nomenclature of the symbolism involved in  this paper.}
	\label{meaning}
	\renewcommand\arraystretch{1.25}
	\begin{tabular}{p{10mm}<{\centering}p{63mm}<{\centering}}
		\hline
		\textbf{Symbolism} & \textbf{Meaning} \\ \hline
		$S_{\rm{hazy}}$ &  Domain set of hazy samples\\
		$S^i$ & The $i$th domain\\
		$D^i$ & The $i$th task\\
		$x$ & Hazy images\\
		$y$ & Haze-free images\\
		$\phi^i$ & Task-specific parameter of $D^i$\\
		$\phi^{i*}$ & Optimal domain-specific parameter of $S^i$\\
		$\varphi$ & Preliminary parameter \\
		$E$& Adaptation network\\
		$F$& Dehazing network\\
		${\omega_1}$& Neural weights of adaptation network\\
		${\omega_2}$& Neural weights of dehazing network\\
		$M$&Number of sample pairs in each task\\
		$N$&Number of sampled tasks\\
		$I$& Number of domains in $S_{\rm{hazy}}$\\
		$K$ & Number of preliminary parameters of a task\\

		\hline

	\end{tabular}
\end{table}

\subsection{Meta-Learning in Image Restoration}
Meta-learning, also known as learning to learn, targets adapting to a new scenario rapidly from a limited number of samples, which has been applied to diverse computer vision tasks and has made significant breakthroughs in recent years.  
Meta-learning is often categorized into metric-based, optimization-based, and model-based techniques \citep{huisman2021survey}, where optimization-based algorithms, especially  model-agnostic meta-learning (MAML) \citep{finn2017model} and its variants \citep{liu2019self,sun2020test},  are widely employed  in image restoration to refine the model parameters at test-time for improving the generalization capability. \cite{soh2020meta} employ MAML to   image super-resolution for obtaining an optimal model initialization, based on which the model can adapt to unseen samples with several test-time training steps. \cite{chi2021test} adopt an auxiliary reconstruction task to optimize the  model indirectly to deal with blur images caused by unseen kernels. To tackle multi-domain learning in image dehazing,  \cite{liu2022towards} conduct test-time training to enable the adaptation to specific domains. 

However, the effectiveness of test-time training depends heavily on manually designed hyperparameters (i.e., iteration steps and learning rates), which may be various among different test samples and lead to under-fitting on target unsen images \citep{gao2022matters}. In addition, test-time training increases the computational costs and runtime of the model \citep{huisman2021survey,liu2022towards}, which results in low efficiency in practical applications. To tackle these issues, this paper resorts to model-based meta-learning approaches \citep{garnelo2018conditional,garnelo2018neural,zhang2021adaptive}, especially ARM \citep{zhang2021adaptive}, to modify the model without sensitive and time-consuming test-time training. Note that although both our work and  that in  \cite{liu2022towards} are based on meta-learning, the problem definition is different. \cite{liu2022towards} require that  test hazy images  are from the same domain as the training ones, while we attempt to overcome the out-of-distribution challenge with test hazy images from unseen  real domains.

\section{Methodology}
\label{section3}

\subsection{Framework Overview}

\begin{figure}[!t]
\centering
\includegraphics[width=\linewidth]{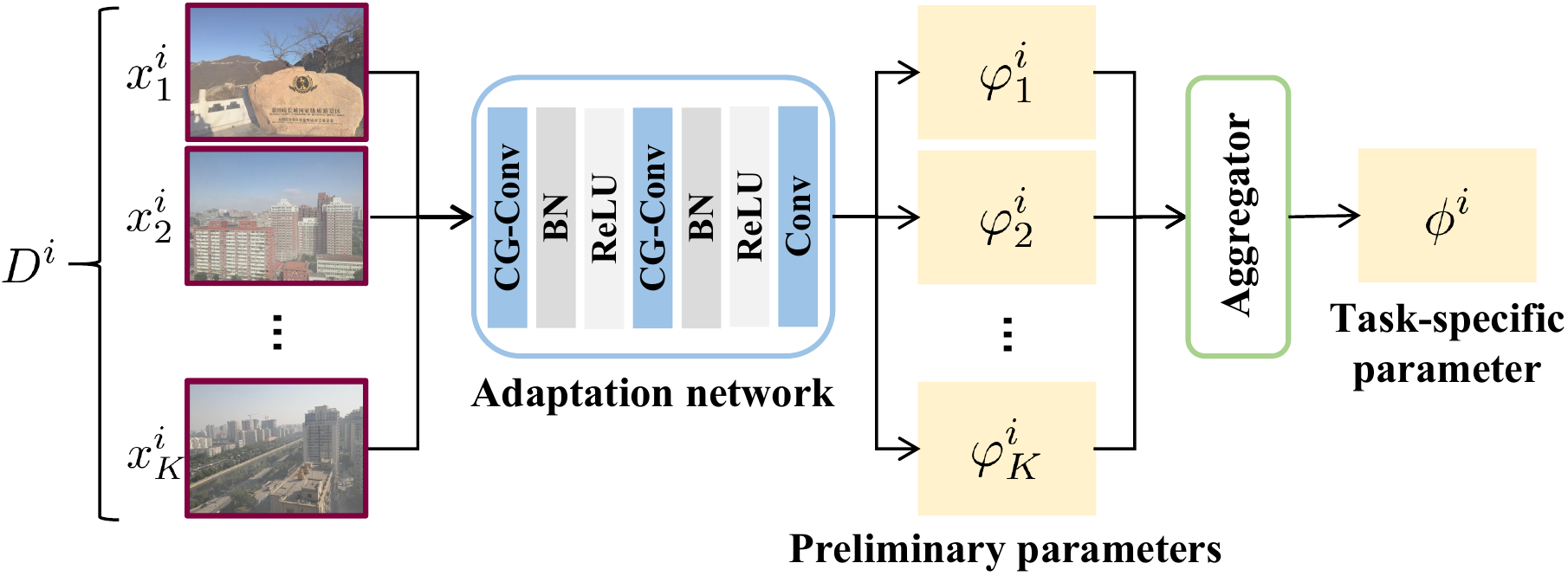}
\caption{The process of obtaining task-specific parameters.}
\label{fig:agre}
\end{figure}

In this paper, we  present a novel framework for single image dehazing, which can deal with out-of-distribution domain generalization  and enable internal learning on real domains without test-time training. 
As shown in Figure~\ref{fig:framework1}, 
our proposed framework includes an adaptation network $E_{\omega_1}(\cdot)$ and a dehazing network $F_{\omega_2}(\cdot,\phi)$.  $\omega_1$ and  $\omega_2$ stand for the neural parameters of $E_{\omega_1}(\cdot)$ and  $F_{\omega_2}(\cdot,\phi)$, respectively, and $\phi$ denotes external variables and serves as an additional input of $F_{\omega_2}(\cdot,\phi)$.  Among them, $\omega_1$ and  $\omega_2$ are fixed after meta-training and are shared cross domains, while  $\phi$  varys with domain properties of input hazy images and is specified to a particular domain.

Assume that there are $I$ domains $S_{\rm{hazy}}=\{S^i\}_{i=1}^I$ with  $M$ hazy  and haze-free pairs $\{(x_k,y_k)\}_{k=1}^M$, and the optimal external variables of  $I$ domains are represented as domain-specific parameters $\{\phi^{i*}\}_{i=1}^I$.  Intuitively, $\{\phi^{i*}\}_{i=1}^I$ embody the most representative and discriminative internal information of corresponding domains.  
Given a specific domain $S^i$ sampled from $S_{\rm{hazy}}$ and a  task $D^i=\{(x_k^i,y_k^i)\}_{k=1}^K$ sampled from $S^i$, our aim is to  estimate a task-specific parameter $\phi^i$ from  $D^i$ to approximate $\phi^{i*}$, so that the regressed dehazing function $F_{\omega_2}(\cdot,\phi^i)$ is capable of handling $M$ samples in $S^i$ commendably. We adopt $E_{\omega_1}(\cdot)$  to dig out internal information related to $S^i$ from each sample in $D^i$ and cache the distilled information into a series of features $\{\varphi_k^i\}_{k=1}^K$, which we call preliminary parameters in this paper.  These preliminary parameters are then aggregated to obtain   $\phi^i$ via a permutation invariant operation  \citep{garnelo2018conditional,garnelo2018neural,zhang2021adaptive,ye2022contrastive}.  In this way, $N$ distinct and powerful dehazing functions $\{F_{\omega_2}(\cdot,\phi^i)\}_{i=1}^N$ can be obtained for a batch of corresponding  tasks $\{D^i\}_{i=1}^N$ without test-time training, where $\{D^i\}_{i=1}^N$ are randomly sampled from various domains in $S_{\rm{hazy}}$. The nomenclature of the symbolism are listed in Table. \ref{meaning}.

\subsection{Adaptation Network}

As exhibited in Figure~\ref{fig:framework1}, $\phi^i$ of $D^i$ is obtained through our adaptation network $E_{\omega_1}(\cdot)$ as well as our aggregator.  
The $E_{\omega_1}(\cdot)$ explores the domain properties from the samples in $D^i$ and store them into a series of  preliminary parameters $\{\varphi_k^i\}_{k=1}^K$.  
The aggregator summarizes the internal information hidden in $\{\varphi_k^i\}_{k=1}^K$  to obtain $\phi^i$. 
In this section, we focus on  our designed $E_{\omega_1}(\cdot)$ and discuss our aggregator in  next section. 

\begin{figure}[t]
\centering
\includegraphics[width=0.85\linewidth]{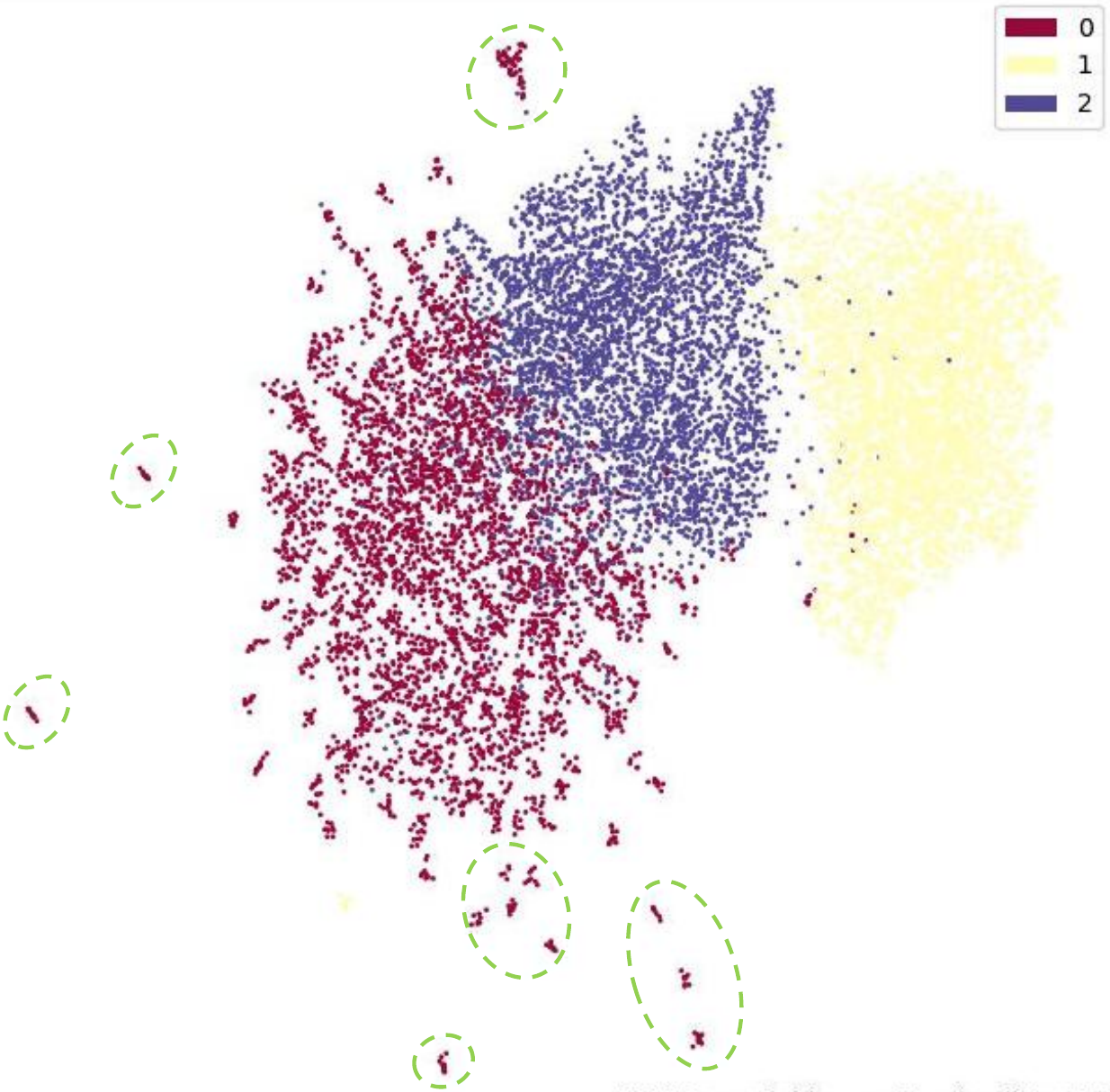}
\caption{UMAP feature visualization of 12000 images randomly sampled from OTS, ITS and RTTS datasets \citep{resideli}, which are denote as the label 0, label 1 and label 2, respectively. The features generated by a pretrained ResNet101 \citep{he2016deep}. Some outliers of the OTS dataset are highlighted by the green dotted circles.}
\label{fig:distribution}
\end{figure}

The architecture of  $E_{\omega_1}(\cdot)$ is displayed in Figure~\ref{fig:agre}.  
Since the internal information of input images is generally covered by haze, directly  employing vanilla convolution layers, which adopt a local perspective to extract domain information,  may lead to  unreliable  $\{\varphi_k^i\}_{k=1}^K$ with poor internal information. 
In this work, CG-Conv \citep{lin2020context} is introduced to compose the principal part of $E_{\omega_1}(\cdot)$. By adopting CG-Conv layers, the extracted internal information  is capable of integrating context information from entire images, so as to access more robust $\{\varphi_k^i\}_{k=1}^K$ with  richer  domain properties. 
In particular, $E_{\omega_1}(\cdot)$ consists of two CG-Conv layers. Each layer is followed by a batch normalization (BN) and a ReLU activation function, where the BN layer is employed to accelerate the convergence of the network. Moreover, a conventional convolution layer is connected to output the preliminary parameter $\varphi_k^i$ for each input sample $x_k^i$. 

\subsection{Distance-Aware Aggregation}

\begin{figure}[!t]
\centering
\includegraphics[width=\linewidth]{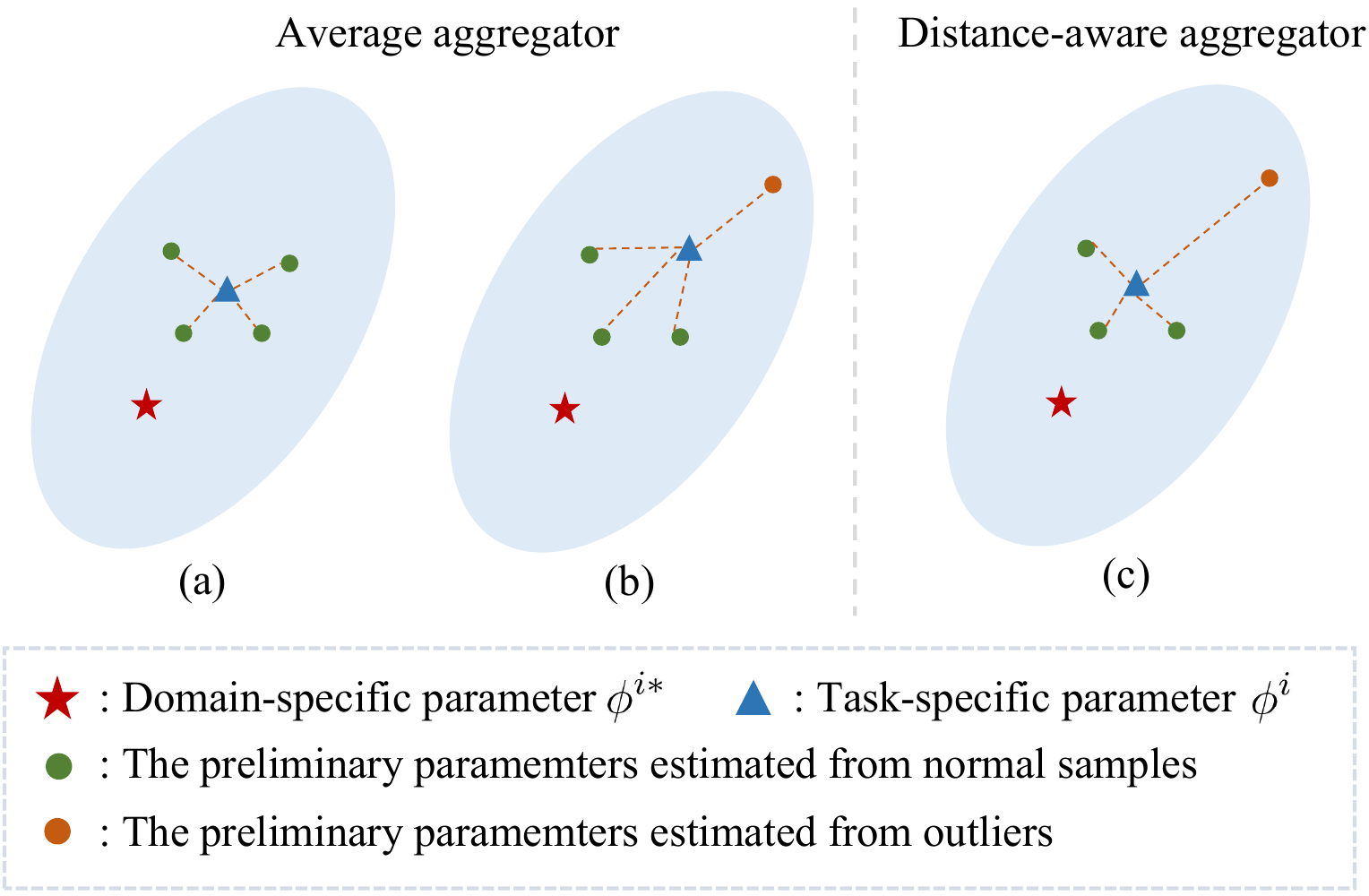}
\caption{Comparison between the average operation and our distance-aware aggregation.  
The blue ellipse represents the distribution of $\{\varphi_k^i\}_{k=1}^M$ estimated by $M$ samples in $S^i$.}
\label{fig:mean}
\end{figure}

Apart from normal samples, there are some outliers in hazy domains.  The  features generated by outliers are usually located far away from those of normal samples, as shown in the green dotted circles  in Figure \ref{fig:distribution}, and fail to capture representative domain-specific internal properties. Thus, it is necessary for the aggregator to suppress misleading information of outliers.
A commonly-used method to aggregate $\{\varphi_k^i\}_{k=1}^K$ for $\phi^i$ is to treat the internal properties extracted from each sample equally and employ an average operation to obtain $\phi^i$  \citep{garnelo2018conditional,garnelo2018neural,zhang2021adaptive,ye2022contrastive}: 
\begin{equation}
\begin{split}
\phi^i=&{1\over K}\sum\limits_{k=1}^K{\varphi_k^i}. 
\end{split}
\end{equation}
However, when encountering outliers,  the average aggregator  may lead $\phi^i$ to deviate from $\phi^{i*}$,  as depicted in Figure~\ref{fig:mean}(a) and (b). 

Considering the feature distribution of outliers and  normal samples,  we propose a non-parametric distance-aware aggregation operation to alleviate the adverse effects caused by outliers. Specifically, we aim to reduce the contributions of outliers  on $\phi^i$ and heighten that of  normal samples. We first calculate the average distance $d_k^i$ between $k$th preliminary parameter $\varphi_k^i$ and  remaining  ones of $D^i$: 
\begin{equation}
\begin{split}
d_k^i= &{{{1}} \over {K-1}}\sum\limits_{s=1,s\neq k}^{K}  {||\varphi_k^i-\varphi_s^i||_1}, 
\end{split}
\end{equation}
where $||\cdot||_1$ denotes the $L1$ regularization. Thus, we can obtain a set of distance values $\{d_k^i\}_{k=1}^K$ for $K$ samples in $D^i$. The larger the value of $d_k^i$, the higher the probability that $x_k^i$ can be regarded as an outlier and the weight coefficient of $\varphi_k^i$ needs to be reduced. 
Then,  we reset the weights of $\{\varphi_k^i\}_k^{K}$  according to the computed $\{d_k^i\}_{k=1}^K$, and   obtain $\phi^i$  by summing the reweighted preliminary parameters:
\begin{equation}
\begin{split}
\phi^i =&\sum\limits_{k = 1}^K {  {{{exp(-d_k^i)}} \over {\sum_{s=1}^K{exp(-d_s^i)}}}\varphi_k^i}.
\end{split}
\end{equation}
In this way, $\phi^i$ can capture more representative internal information from normal samples and less misleading information from outliers, which encourages  $\phi^i$ to be located closer to $\phi^{i*}$, as illustrated in Figure \ref{fig:mean}(c).

\subsection{Domain-Relevant Contrastive Regularization}
\begin{figure}[!t]
\centering
\includegraphics[width=0.95\linewidth]{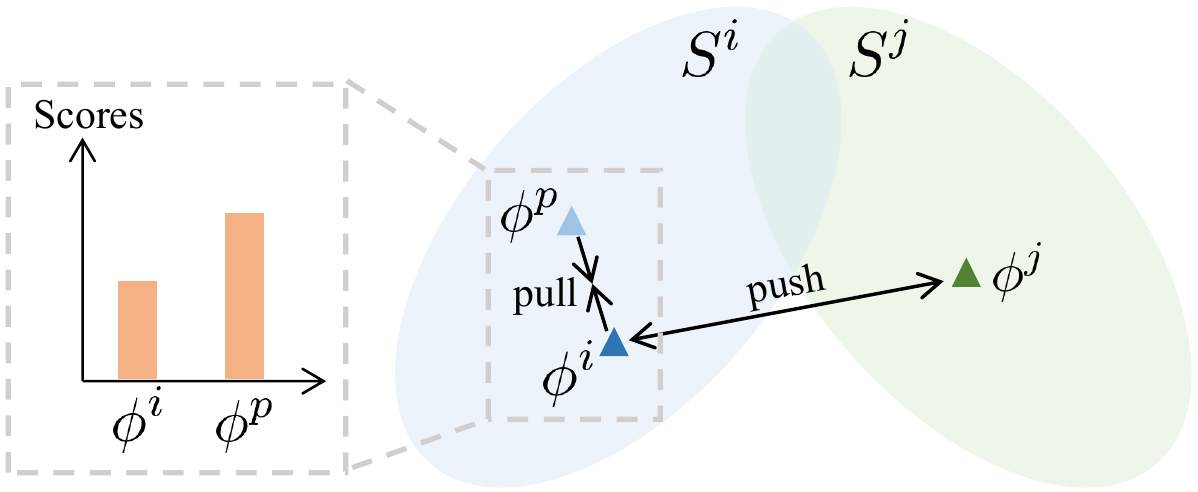}
\caption{An example to illustrate the domain-relevant contrastive regularization, where $\phi^i$ and $\phi^p$ are from $S^i$ and $\phi^j$ are from $S^j$.  }
\label{fig:score}
\end{figure}

To regress a more discriminative and representative task-specific parameter, in this paper, we incorporate inter-domain information with intra-domain properties. 
We note that each task-specific parameter in $\{\phi^i\}_{i=1}^N$ is a linear combination of corresponding preliminary parameters and  can be regarded as a collection of internal features. It is reasonable to assume that the task-specific parameters generated from the same domain have more similar internal features, while the ones  from different domains have more distinct features.  Based on this assumption, we introduce contrastive learning \citep{he2020momentum,chen2020simple}  and propose a novel domain-relevant contrastive regularization to capture the intra-domain homogeneity and inter-domain heterogeneity.  Figure \ref{fig:score} provide an example to depict our domain-relevant contrastive regularization.

For the sake of measuring the feature similarities of  task-specific parameters  generated from non-aligned tasks, we draw lessons from some studies in image-to-image translation \citep{zhang2020cross,zhan2021unbalanced} and employ the contextual loss $L_{cx}$ via: 
\begin{equation}
\begin{split}
L_{cx}(\phi^{i},\phi^{j}) = &-log(\sum\limits_{u}  {\mathop{max}\limits_{v}A_{uv}(\phi^{i,u},\phi^{j,v})}),  
\label{contextual_loss}
\end{split}
\end{equation}
where $u$ and $v$ are the indexes of the feature maps in  $\phi^i$ of  $D^i$  and  $\phi^j$ of  $D^j$, respectively. $A_{uv}$ is the contextual similarity  measurement commonly adopted  in image-to-image translation \citep{mechrez2018contextual}. The remaining question is how to choose ``positive'' and ``negative'' pairs.

Suppose that merely $\phi^i$ and $\phi^p$ in $\{\phi^i\}_{i=1}^N$  are from the same domain $S^i$, while the others are from diverse domains except $S^i$. Our idea for the ``positive'' pair is to pick the more representative one from $\phi^i$ and $\phi^p$ and adopt  it to guide another to capture more representative internal information. Targeting this issue, we embed an additional classifier into our framework, as exhibited in Figure \ref{fig:framework1}, which takes the  $\{\phi^i\}_{i=1}^N$  as inputs and attempt to predict the confidence scores. The higher the confidence score, the more representative the parameter is and easier it to be classified into the corresponding domain. If the confidence score of $\phi^p$ is higher than that of $\phi^i$, $\phi^p$ will be selected to assist $\phi^i$ to learn more representative internal information, and vice versa. For the ``negative'' pairs,  the unpicked parameter is grouped with each task-specific parameter generated from other domains to enhance its discrimination. 
Therefore, our domain-relevant contrastive regularization $L_{DCR}$ can be deduced as: 
\begin{equation}
\begin{split}
L_{DCR}=&{{L_{cx}(\phi^i,\phi^p)} \over {{L_{cx}(\phi^i,\phi^p)+\sum\limits_{s = 1,s\neq i,s\neq p}^N {L_{cx}(\phi^i,\phi^s)}+\sigma}}},
\end{split}
\end{equation}
where the picked parameter is $\phi^p$ that has higher confidence scores than $\phi^i$. The $\sigma$ is a constant to avoid situations where the denominator becomes zero.

\subsection{Loss Function}
Except for the domain-relevant contrastive regularization $L_{DCR}$,  there are other four loss functions  are employed in our experiments, including the pixel-wise loss $L_{pixel}$ \citep{dong2020fd,qin2020ffa}, the structure similarity loss $L_{SSIM}$ \citep{dong2020fd}, the contrastive regularization loss $L_{CR}$ \citep{wu2021contrastive} and the cross entropy loss $L_{CE}$.  Therefore, the overall optimization function $L$ of a batch of $n$ sampled tasks is defined as 
\begin{equation}
\begin{split}
L=&L_{pixel}+\lambda_1 L_{SSIM} + \lambda_2 L_{CR} + \lambda_3 L_{CE} +  \lambda_4 L_{DCR},
\end{split}
\end{equation}
where $\lambda_1$, $\lambda_2$, $\lambda_3$ and $\lambda_4$  are the trade-off weights. 

\subsubsection{Pixel-wise Loss.}
Pixel-wise loss is employed to quantify the pixel-wise distance between the generated image and the ground-truth: 
\begin{equation}
\begin{split}
L_{pixel}=&{{1} \over {N \times K}}\sum\limits_{i= 1}^N{\sum\limits_{k = 1}^K{||\hat{y}_k^i-y_k^i||_1}},  
\end{split}
\end{equation}
where the $\hat{y}_k^i$ is the haze-free image generated from the $k$th sample $x_k^i$ in the $i$-th task.

\subsubsection{Structure Similarity Loss.}
Structure similarity loss $L_{SSIM}$ quantifies the distance between the restored image and the ground truth in terms of brightness and contrast. It is defined as
\begin{equation}
\begin{split}
L_{SSIM}=& {{1} \over {N \times K}}\sum\limits_{i= 1}^N{\sum\limits_{k = 1}^K{(1-SSIM(\hat{y}_k^i,y_k^i))}},
\end{split}
\end{equation}
where $SSIM(\cdot,\cdot)$ stands for the operation to calculate the structure similarity of two images.

\subsubsection{Contrastive Regularization Loss.}
We also employ the contrastive regularization \citep{wu2021contrastive} to further meliorate the quality of the restored images in the representative space. 
\begin{equation}
\begin{split}
L_{CR}=& {{1} \over {N\times K}}\sum\limits_{i= 1}^N{\sum\limits_{k = 1}^K{\sum\limits_{s = 1}^T \alpha_s \frac {|| V_s(y_k^i),V(\hat{y}_k^i)||_1 } { || V_s(x_k^i),V(\hat{y}_k^i)||_1}}}, 
\end{split}
\end{equation}
where $V(\cdot)$ denotes the fixed pre-trained feature extractor and $V_i(\cdot)$ is $s$-th feature map from the feature extractor. $T$ is the number of feature maps. $\beta_s$ is the balancing term of the contrastive regularization loss that relates to the $s$-th feature map. 
\subsubsection{Cross Entropy Loss.}
Cross entropy loss $L_{CE}$ is employed to train our classifier, which is defined as:
\begin{equation}
\begin{split}
L_{CE}=&-{{1} \over {N}}\sum\limits_{i = 1}^N {P(\phi^i)\log_e(Q(\phi^i))},
\end{split}
\end{equation}
where the $P(\phi^i)$ and $Q(\phi^i)$ are the given probability and the estimated probability of $\phi^i$, respectively. The $e$ is the Euler number. 

\section{Experiments}
\label{section4}

\begin{table*}[width=2.08\linewidth,cols=4,pos=h]
\renewcommand{\arraystretch}{1.25}
\begin{threeparttable}[b]
\caption{Quantitative comparison  of the SOTA dehazing models on real hazy datasets \citep{resideli,ancuti2018haze,ancuti2020nh}.}
\label{table:dehazing}
\begin{tabular*}{\tblwidth}{p{16mm}<{\centering}|p{7mm}<{\centering}|p{7mm}<{\centering}|p{12mm}<{\centering}p{12mm}<{\centering}|p{12mm}<{\centering}p{12mm}<{\centering}|p{12mm}<{\centering}p{12mm}<{\centering}|p{12mm}<{\centering}|p{13mm}<{\centering}}
	\hline
	\multirow{2}{*}{Methods}  & \multirow{2}{*}{Year} & \multirow{2}{*}{Real} & \multicolumn{2}{c|}{RTTS} & \multicolumn{2}{c|}{URHI}  & \multicolumn{2}{c|}{ESPW} & \multirow{2}{*}{\#Param}& \multirow{2}{*}{Runtime(s)}    \\
	& &&  BRISQUE$\downarrow$ & NIQE$\downarrow$  &  BRISQUE$\downarrow$& NIQE$\downarrow$ &   BRISQUE$\downarrow$& NIQE$\downarrow$ &&   \\ \hline
	Hazy &- &-&  37.011* &	3.583*&	33.531&	4.128&21.081&2.859&-&- \\  \rowcolor{gray!10}
	DAD&2020&$\surd$&32.727*&3.672*&-&-&29.728&3.439&54.59M&0.010\\ \rowcolor{gray!10}
	PSD&2021&$\surd$&25.239*&3.077*&-&-&23.616&3.259&33.11M&0.024\\ 
	AOD-Net  &2017&& 35.466	&3.636&	34.077&	3.605 &21.911&3.288&\textbf{0.002M}& \textbf{0.004}\\ 
	GDN  &2019& &28.086&	3.200&	27.941&	4.924 &20.613&\textbf{2.861}&0.96M& 0.014\\ 
	MSBDN  &2020&& 28.743*	&3.154*		&26.617	&3.079&	22.352&2.939&31.35M&0.021  \\ 
	FFA-Net  &2020&& 30.183&	3.050&	26.141&	3.265 &22.753&2.959&4.68M& 0.087\\ 
	AECR-Net &2021&& 28.594&	3.139&	25.879&	3.251 &17.840&2.964&2.61M&0.028 \\ 
	4kD  &2021 && 27.254&	3.149&	24.853&	4.975 &19.116&3.085&34.55M& 0.094 \\ 
	D4  &2022&& 29.536&	3.174&	27.429&	3.065 &18.987&2.906&10.70M& 0.032\\ 
	DeHamer &2022&& 30.986&	3.197&	28.202&	3.023 &20.723&2.889&4.63M& 0.066  \\ 
	Ours &2022 && \textbf{26.521}&	\textbf{2.997	}&	\textbf{24.465}&	\textbf{2.941}&\textbf{17.726}&2.887&31.58M& 0.062	 \\   
	\hline
\end{tabular*}
\begin{tablenotes}
	\item[1]  ``Real'' denotes the access of real hazy images during the training stage.
	\item[2]  ``*''  represents that the results are obtained from the existing paper \citep{chen2021psd}. 
\end{tablenotes}
\end{threeparttable}
\end{table*}

\begin{figure*}
\centering
\includegraphics[width=\linewidth]{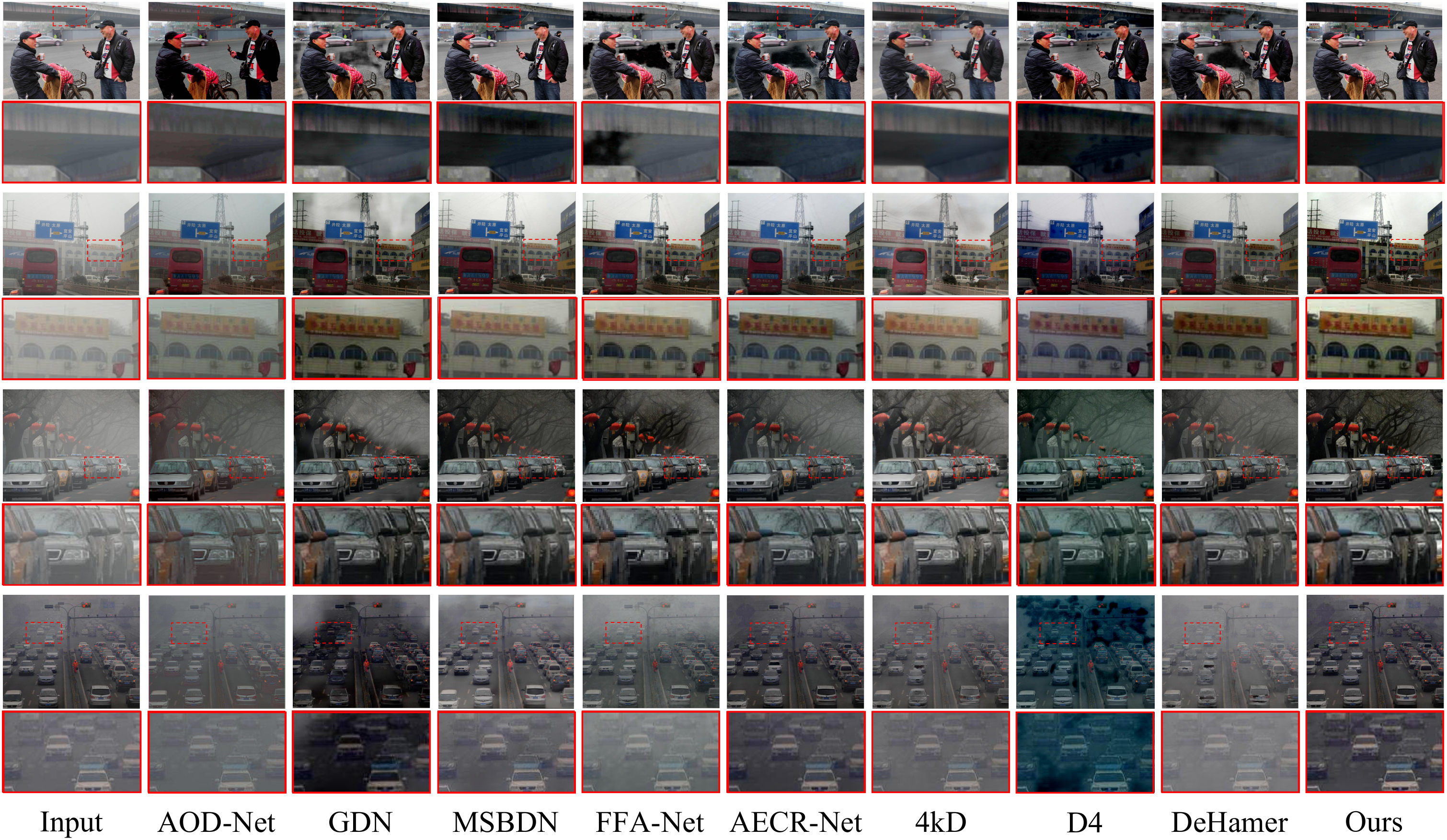}
\caption{Visual comparisons with conventional learning-based methods on real-world hazy images from RTTS dataset \citep{resideli}.}
\label{fig:rtts}
\end{figure*}

\subsection{Implementation Details}

\subsubsection{Datasets} RESIDE \citep{resideli} is  a widely-used dataset for single image dehazing, which consists of five subsets, including Indoor Training Set (ITS), Outdoor Training Set (OTS), Synthetic Object Testing Set (SOTS), Real Task-driven Testing Set (RTTS) and Unannotated Real Hazy Images (URHI). Among them, ITS, OTS, and SOTS are  synthesized by   adjusting the values of scattering coefficient  and atmospheric light artificially, while  RTTS and  URHI are taken in real hazy scenarios directly.  
In our experiments, we select 6000 pairs of images from ITS and OTS, respectively, and create our training set that is composed of 12000 pairs of synthetic samples. We define each synthetic set as a particular domain, as the average depth errors are different among diverse datasets \citep{resideli}.  Therefore, our training set can be structured into two domains.  To evaluate the proposed method on real hazy images, RTTS and URHI are adopted in our experiments, where RTTS is composed of 4322  real hazy images and URHI consists of 4809 ones. We also evaluate our model on other real hazy images employed by some previous work \citep{fattal2014dehazing,he2010single}, which are denoted as ESPW in this paper.

\subsubsection{Training Details}  In our experiments, the coefficients $\lambda_1$, $\lambda_2$, $\lambda_3$  and $\lambda_4$ are set to 0.5, 0.1, 1 and 0.5, respectively, to balance the value of each loss function. The $\sigma$ is set to $10^{-7}$. The feature extractor employed for the contrastive regularization loss $L_{CR}$ is the frozen pre-trained VGG19, and  the  features are selected from the $1$st, $3$rd, $5$th, $9$th and $13$th layers of the feature extractor with the  coefficients $\alpha_s$  set to ${1 / 32},{1 / 16},{1 /8},{1 / 4}$ and $1 $, respectively \citep{wu2021contrastive}. Our proposed model is implemented on the basis of PyTorch, and is trained by the Adam optimizer with $\beta_1=0.9$ and $\beta_2=0.999$.  In each iteration, the model samples 2 tasks, and each task  consists of 4 synthetic hazy images from the same domain.  The initial learning rate is set as 0.0002 for the whole network. In addition, the input size is $240\times240\times3$. The training data is augmented by means of random rotation and random flip. For the network architecture, we follow the previous work MSBDN \citep{dong2020multi} due to its compact architecture and effective performance. All experiments are conducted on a NVIDIA Tesla V100 GPU. 

\begin{figure*}
\centering
\includegraphics[width=\linewidth]{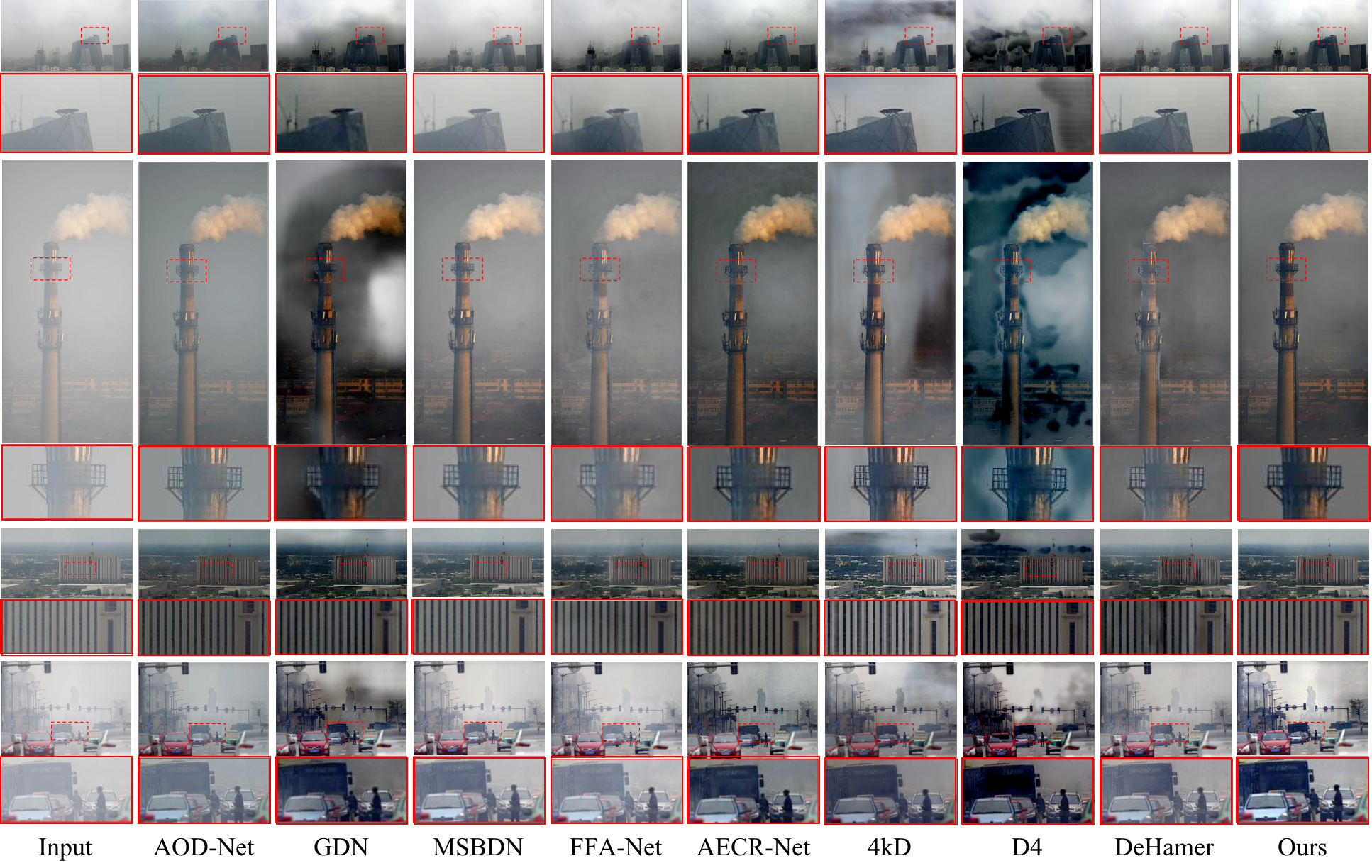}
\caption{Visual comparisons  with conventional learning-based methods  on real-world hazy images from URHI dataset \citep{resideli}.}
\label{fig:urhi}
\end{figure*}

\begin{figure*}
\centering
\includegraphics[width=\linewidth]{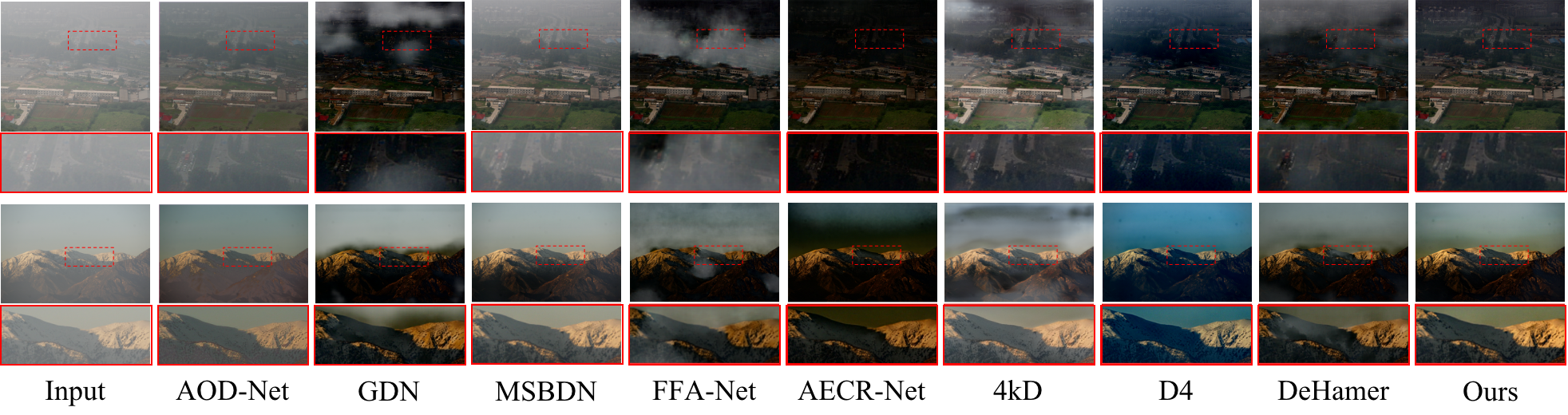}
\caption{Visual comparisons  with conventional learning-based methods  on real hazy images from ESPW dataset \citep{fattal2014dehazing,he2010single}.}
\label{fig:color}
\end{figure*}

\subsubsection{Competitors and Evaluation Metrics} Our proposed model is compared with open-source and state-of-the-art learning-based algorithms, such as AOD-Net \citep{li2017aod},  GridDehazeNet (GDN) \citep{liu2019griddehazenet}, MSBDN \citep{dong2020multi}, FFA-Net \citep{qin2020ffa}, AECR-Net \citep{wu2021contrastive}, 4kDehazing (4kD) \citep{zheng2021ultra}, D4 \citep{yang2022self} and  DeHamer \citep{guo2022image}.  Some image dehazing models based on domain adaptation, which have been  pre-trained or fine-tuned on a large number of real hazy images, are also involved to assess our proposed model, such as DAD \citep{shao2020domain} and PSD \citep{chen2021psd}. 
The results of the competitors are from existing papers, if available. Otherwise, the results are generated through the pre-trained models provided by their authors.

Due to  the absence of paired samples, commonly-used structural similarity index  (SSIM) and  peak signal to noise ratio (PSNR) fail to be applied for the assessment on the dehazing results  in RTTS, URHI \citep{resideli} and ESPW \citep{fattal2014dehazing,he2010single}. Thus, we resort to blind evaluation metrics for providing quantitative comparison on real hazy images. In particular,  we adopt blind/referenceless image spatial quality evaluator (BRISQUE) and natural image quality evaluator (NIQE) in the following experiments. The lower values of BRISQUE and NIQE, the higher quality of restored images.

\subsection{Comparison with State-of-The-Art Methods}

\begin{figure*}
\centering
\includegraphics[width=0.95\linewidth]{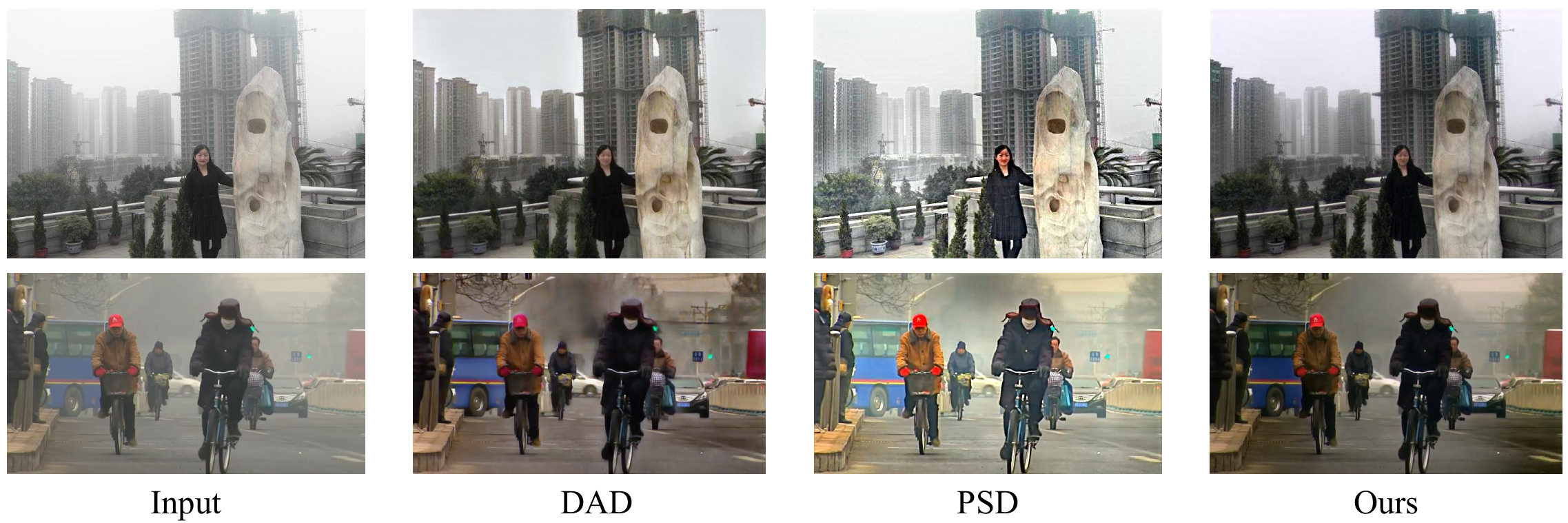}
\caption{Visual comparisons  with domain adaptation-based methods on real hazy images \citep{resideli}.}
\label{fig:adaptation}
\end{figure*}

\begin{figure*}
\centering
\includegraphics[width=\linewidth]{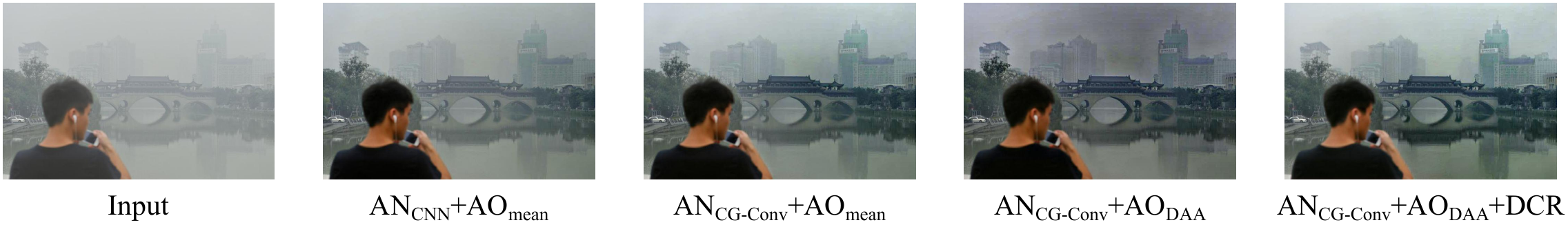}
\caption{Ablation study of our proposed model with different settings on the RTTS dataset \citep{resideli}.}
\label{fig:ablation}
\end{figure*}

\subsubsection{Quantitative Comparison} We first leverage BRISQUE and NIQE to evaluate the performance of our proposed model with the state-of-the-art competitors. Table \ref{table:dehazing} reveals the quantitative  results of each participant on   RTTS, URHI \citep{resideli} and ESPW \citep{fattal2014dehazing,he2010single} datasets.  The best results of dehazing models trained merely on synthetic samples are shown in bold, and the runtime of each method is  obtained by averaging the  cost of  1000 images with the size of $480 \times 640$. Due to the  exposure to URHI dataset during training or fine-tuning, DAD \citep{shao2020domain} and PSD \citep{chen2021psd} are only evaluated on RTTS and ESPW datasets. 
It can be clearly observed that our proposed model has the better performance against the other conventional learning-based algorithms in the case of training on synthetic data. In addition, our model is competitive to the domain adaptation-based algorithms, which have the access to real hazy images before testing.  
Compared with our backbone MSBDN \citep{dong2020multi}, our model  achieves considerable performance gains on real hazy images with a small amount of additional parameters and runtime costs.

\begin{table}[!t]
\renewcommand{\arraystretch}{1.25}
\centering
\caption{Quantitative comparison  of object detection on RTTS dataset \citep{resideli}.}
\label{table:detection}
\begin{tabular}{p{19mm}<{\centering}|p{6mm}<{\centering}|p{20mm}<{\centering}|p{20mm}<{\centering}}
\hline
Methods & Real  & mAP (\%) & Gain   \\ \hline
Hazy& -&63.32&	-\\  \rowcolor{gray!10}
DAD&$\surd$&65.02&+1.70\\  \rowcolor{gray!10}
PSD&$\surd$&65.84&+2.52\\ 
AOD-Net &&60.45&	-2.87\\
GDN &&63.59&	+0.27  \\
MSBDN   &&65.16&	+1.84  \\
FFA-Net  &&64.44&	+1.12  \\
AECR-Net  &&65.39&	+2.07 \\
4kD  &&64.53&	+1.21  \\
D4& &63.44&	+0.12  \\
DeHamer&&64.63 &+1.31\\
Ours &&\textbf{65.55}	&\textbf{+2.23}  \\  
\hline
\end{tabular}
\end{table}

\subsubsection{Qualitative Comparison} In this section, we provide qualitative results to further evaluate the performance of our proposed model. Figure \ref{fig:rtts}, Figure \ref{fig:urhi} and Figure \ref{fig:color}  exhibit the visual comparison  of both our model and conventional learning-based algorithms on RTTS, URHI and ESPW datasets \citep{resideli}, respectively. It can be seen that our proposed model can obtain higher-quality dehazing results from both global  and local perspectives, where the images with less color distortion and more detailed textures can be restored. Our model is improved on  MSBDN \citep{dong2020multi} but is capable of restoring more clear objects  with less residual haze, which demonstrates that utilizing the internal information of real domains contributes to boost the model performance on real hazy images. Compared with our model, there are more challenges for the competitive algorithms to deal with thick haze and color shift in hazy inputs, leading to erroneously-handling areas and darker visual perception of their restored images. We also exhibit the dehazing results of the algorithms based on domain adaptation in Figure \ref{fig:adaptation}. It can be clearly found that although our model fail to be pre-trained and fine-tuned on real hazy images, the restored image of our model has less residual haze than that of PSD \citep{chen2021psd} and has visual effects similar to that of DAD \citep{shao2020domain}.

\begin{table*}[width=2.08\linewidth,cols=4,pos=h]
\renewcommand{\arraystretch}{1.25}
\centering
\caption{Ablation study of our proposed model with different settings on  URHI dataset \citep{resideli}.  }
\label{table:ablation1}
\begin{tabular*}{\tblwidth}{p{23mm}<{\centering}|p{13mm}<{\centering}p{13mm}<{\centering}p{13mm}<{\centering}p{13mm}<{\centering}p{13mm}<{\centering}p{13mm}<{\centering}|p{17mm}<{\centering}p{17mm}<{\centering}}
\hline
\multirow{3}{*}{Methods} &\multicolumn{6}{c|}{Settings}  &\multicolumn{2}{c}{\multirow{2}{*}{URHI}}  \\ \cline{2-7} 
&\multicolumn{1}{c|}{\multirow{2}{*}{Baseline}}  & \multicolumn{2}{c|}{Adaptation network} & \multicolumn{2}{c|}{Aggregator} &\multirow{2}{*}{DCR}  &&\\
&\multicolumn{1}{c|}{}&$AN_{CNN}$&\multicolumn{1}{c|}{$AN_{CG-Conv}$}&$AO_{mean}$&\multicolumn{1}{c|}{$AO_{DAA}$}& &BRISQUE$\downarrow$&NIQE$\downarrow$ \\ \hline 
\multirow{5}{*}{Our model}&$\checkmark$ && & & &&26.537&2.998\\ 
&$\checkmark$& $\checkmark$&&$\checkmark$ & & &26.470&2.983 \\ 
&$\checkmark$&&$\checkmark$ &$\checkmark$ & & &26.012&2.964 \\ 
&$\checkmark$&&$\checkmark$ & &$\checkmark$ & &25.791&2.955 \\ 
&$\checkmark$&&$\checkmark$ & & $\checkmark$& $\checkmark$&	\textbf{24.465}&	\textbf{2.941} \\ 
\hline
\end{tabular*}
\end{table*}

\begin{figure*}
\centering
\includegraphics[width=0.95\linewidth]{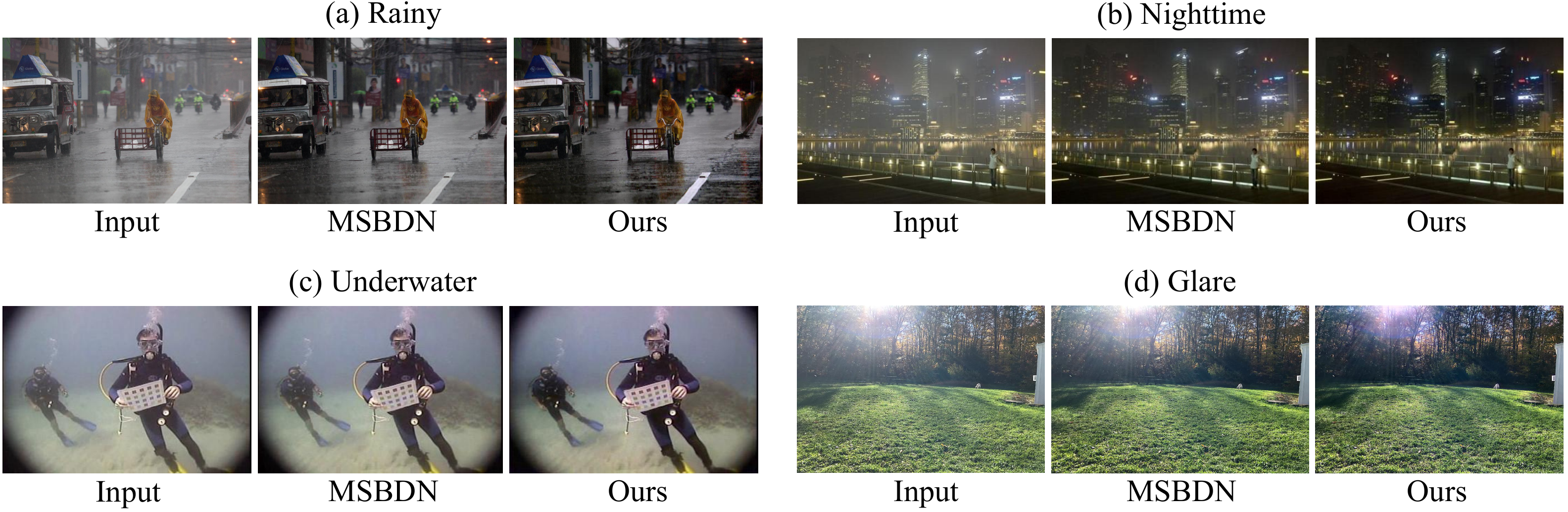}
\caption{Visual comparisons on other similar tasks. The inputs of (a) and (d) are download from the  Internet, and the inputs of (b) and (c) are from the relevant existing datasets of nighttime image dehazing \citep{li2015nighttime,zhang2017fast,zhang2020nighttime} and underwater image enhancement \citep{islam2020fast}.}
\label{fig:other}
\end{figure*}

\subsubsection{Evaluation on Object Detection} 
We further evaluate the generalization ability of the dehazing algorithms according to the performance  improvement of the object detection model. We adopt RTTS dataset \citep{resideli}, which is composed of  real hazy image with annotations of object categories as well as bounding boxes. We use different dehazing algorithms to restore and enhance the images of RTTS \citep{resideli} separately for meliorating the quality of  test samples. We leverage YOLOv3 \citep{redmon2018yolov3} to detect  objects on generated haze-free results, and calculate their mean average accuracy (mAP). Table \ref{table:detection} shows the detection accuracy and performance gains of YOLOv3 \citep{redmon2018yolov3} on diverse dehazing results. It can be noticed that the mAP has a significant decline on the images restored by AOD-Net \citep{li2017aod}, suggesting that AOD-Net \citep{li2017aod} tend to entail   the performance degradation of the object detection model. Moreover, compared with remaining learning-based competitors, the mAP value calculated from our model reaches the top accuracy,  which implies that the  images generated by our model  have higher perceptual sensitivity. Furthermore, in terms of the evaluation driven by  the downstream visual task,  our model is comparable to the domain adaptation-based methods \citep{shao2020domain,chen2021psd}. 

\subsection{Ablation Study}
In this section, we conduct comprehensive ablation studies on the RTTS dataset \citep{resideli} to illustrate the effectiveness of different elements in our proposed model. Figure \ref{fig:ablation} and Table \ref{table:ablation1} exhibit the qualitative and quantitative results generated by our proposed model with diverse settings, respectively. Taking the differences of loss functions into account, we  retrain the MSBDN \citep{dong2020multi} with our loss functions  (denoted as baseline),  so that  the evaluation interference caused by loss functions can be eliminated. $AN_{CNN}$ and $AN_{CG-Conv}$ stands for the adaptation network based on conventional CNN and CG-Conv, respectively. $AO_{mean}$ and $AO_{DAA}$ denotes the the average and the distance-aware aggregation, respectively.  \emph{DCR} is the domain-relevant contrastive regularization. It can be clearly observed in Figure \ref{fig:ablation}  that merely adopting the adaptation network and the average aggregator contributes to residual haze in the restored image. The presented distance-aware aggregator significantly alleviates the residual-haze effect but darkens generated images, leading to poor visual perception. Instead, the combination of the proposed adaptation network, the distance-aware aggregator and the domain-relevant contrastive regularization can produce clearer and higher-quality images. Table \ref{table:ablation1} also illustrates the effectiveness of our proposed elements.

\subsection{Application to Other Low-Level Tasks}
Apart from single image dehazing, we explore some other low-level computer vision tasks, including single image deraining, nighttime image dehazing, underwater image enhancement, and single image glare removal, to evaluate the performance gains of our model compared with the backbone MSBDN \citep{dong2020multi}.  Figure \ref{fig:other} shows the results generated by MSBDN \citep{dong2020multi} and our model. 
It can be seen that our contributions on MSBDN \citep{dong2020multi} not only improve the performance on real hazy samples, but also enhance the generalization capability   in other similar low-level tasks.

\section{Conclusions}
\label{section5}
In this work, we propose a domain generalization framework via model-based meta-learning for single image dehazing. By combining both our  adaptation network and distance-aware aggregator with the dehazing network, our model 
can dig out representative  internal information from a specific real domain. In addition, we present a domain-relevant contrastive regularization to facilitate external variables to capture more  discriminative information of domains, contributing to a more powerful dehazing function for the given domain. The extensive experiments demonstrate that our proposed model outperforms the state-of-the-art competitors on real hazy domains.

\bibliographystyle{cas-model2-names}

\bibliography{cas-refs}

\end{document}